\documentclass[11pt]{article}

\usepackage[preprint]{acl}

\usepackage{times}
\usepackage{latexsym}

\usepackage[T1]{fontenc}

\usepackage[utf8]{inputenc}

\usepackage{microtype}

\usepackage{inconsolata}

\usepackage{graphicx}

\usepackage{hyperref}
\usepackage{url}

\usepackage{amsmath}
\usepackage{graphicx}
\usepackage{booktabs}
\usepackage{geometry}
\usepackage{algorithm}

\usepackage{algpseudocode}
\usepackage{listings}
\usepackage{xcolor}

\lstdefinestyle{trajectory}{
    basicstyle=\footnotesize\ttfamily,
    breaklines=true,
    frame=single,
    framerule=0.5pt,
    backgroundcolor=\color{gray!5},
    rulecolor=\color{gray!30},
    captionpos=b,
    aboveskip=10pt,
    belowskip=10pt,
    xleftmargin=5pt,
    xrightmargin=5pt,
    showstringspaces=false,
    tabsize=2,
    columns=flexible,
    keepspaces=true
}
\usepackage{abstract}
\usepackage{amsthm}
\usepackage{multirow}
\usepackage{wrapfig}
\usepackage{tikz}
\usepackage{fontawesome5} 
\usetikzlibrary{arrows.meta,positioning,calc,decorations.pathreplacing,shadows,shapes.misc,shapes.geometric}
\usepackage{booktabs}
\usepackage{siunitx}
\usepackage{multirow}
\usepackage[table]{xcolor}
\usepackage{svg-extract}
\usepackage{adjustbox}
\usepackage[most]{tcolorbox}
\usepackage{enumitem}
\usepackage{amssymb}
\usepackage{booktabs}
\usepackage{xcolor}
\usepackage[most]{tcolorbox}
\tcbset{
  colback=gray!5!white,
  colframe=gray!75!black,
  boxrule=0.5pt,
  arc=2pt,
  left=1mm,
  right=1mm,
  top=1mm,
  bottom=1mm,
  fontupper=\ttfamily\small,
  enhanced,
  breakable
}
\usepackage{authblk}  

\title{COMPASS: Enhancing Agent Long-Horizon Reasoning with Evolving Context}

\author[1,2]{Guangya Wan\thanks{Work done during internship at Google.}}
\author[1]{Mingyang Ling}
\author[1]{Xiaoqi Ren}
\author[1]{Rujun Han}
\author[2]{Sheng Li}
\author[1]{Zizhao Zhang}

\affil[1]{Google Cloud AI}
\affil[2]{University of Virginia}

\begin{document}

\maketitle
\begin{abstract}
\noindent Long-horizon tasks that require sustained reasoning and multiple tool interactions remain challenging for LLM agents: small errors compound across steps, and even state-of-the-art models often hallucinate or lose coherence. We identify \emph{context management} as the central bottleneck—extended histories cause agents to overlook critical evidence or become distracted by irrelevant information, thus failing to replan or reflect from previous mistakes. To address this, we propose \textbf{COMPASS} (\emph{Context-Organized Multi-Agent Planning and Strategy System}), a lightweight hierarchical framework that separates tactical execution, strategic oversight, and context organization into three specialized components: (1) a \emph{Main Agent} that performs reasoning and tool use, (2) a \emph{Meta-Thinker} that monitors progress and issues strategic interventions, and (3) a \emph{Context Manager} that maintains concise, relevant progress briefs for different reasoning stages. Across three challenging benchmarks—GAIA, BrowseComp, and Humanity’s Last Exam—COMPASS improves accuracy by up to 20\% relative to both single- and multi-agent baselines. We further introduce a test-time scaling extension that elevates performance to match established DeepResearch agents, and a post-training pipeline that delegates context management to smaller models for enhanced efficiency.
\end{abstract}

\section{Introduction}
\label{sec:introduction}

Large Language Model (LLM) agents have achieved impressive performance on tasks with well-defined reasoning paths and objectives \citep{comanici2025gemini25, minaee2025largelanguagemodelssurvey}. The emerging challenge for autonomous systems is mastering \emph{long-horizon tasks} (LHT)—problems that demand sustained reasoning across multiple tool interactions while maintaining strategic coherence and adapting to unexpected outcomes \citep{sun2023adaplanner, xi2025agentgymrltrainingllmagents}. For example, a query from the BrowseComp dataset \citep{wei2025browsecompsimplechallengingbenchmark} may ask to identify a soccer player with specific yellow-card patterns across halves in a given year, requiring database lookups, referee validation, and integration of timing and match data.

\begin{figure*}[t]
\centering
\includegraphics[width=0.89\linewidth]{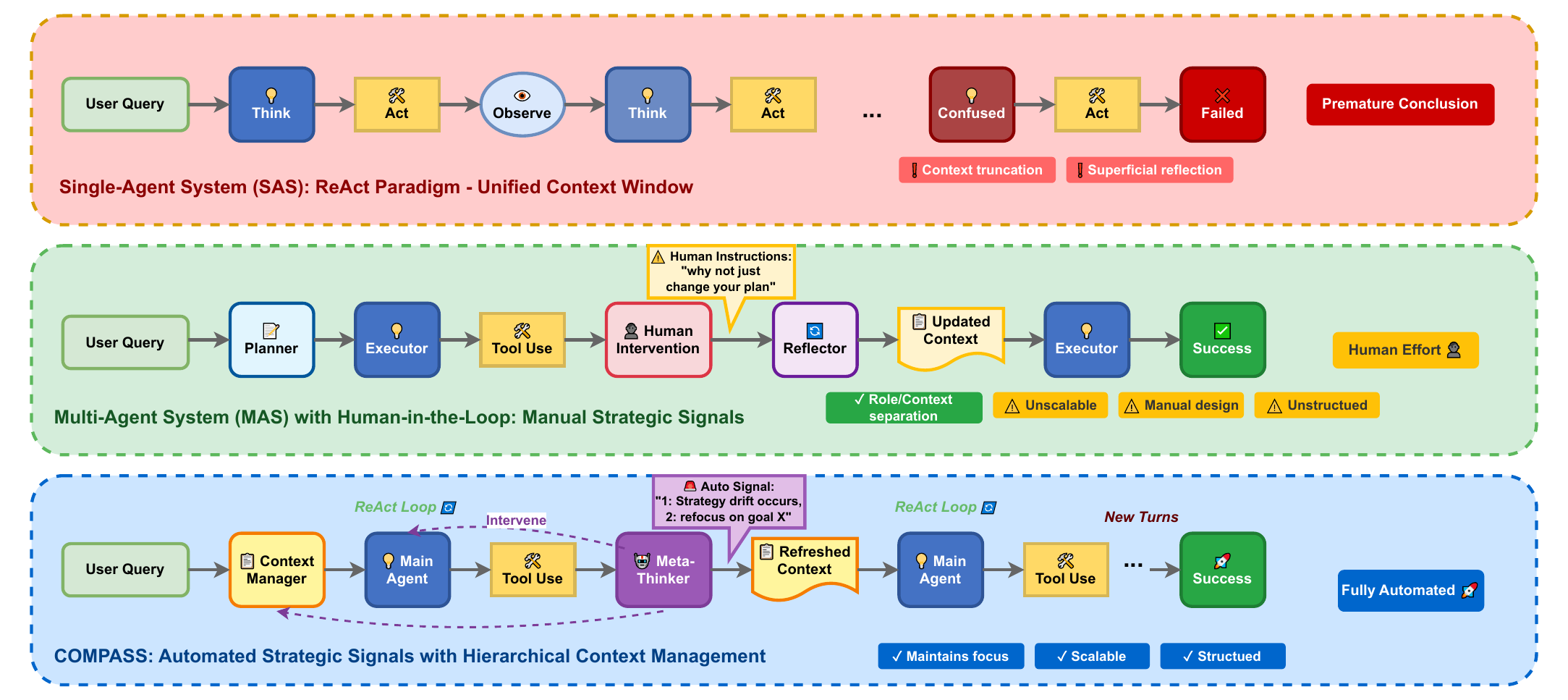}
\caption{\textbf{Motivation for COMPASS.} ReAct-style SAS accumulate full dialogue histories, leading to context exhaustion and performance degradation. MAS with Human-in-the-loop improves performance through human's feedback but require human effort and lack scalability. \textsc{COMPASS} introduces automated monitor and context management agents that track reasoning, organize structured context, and maintain reliability with full automation.}
\vspace{-8pt}
\label{fig:agent_demo}
\end{figure*}

Such tasks are challenging because agents must maintain a high success rate at \textbf{each step} of tool use. Minor errors—such as ambiguous search results or faulty API calls—can cascade, turning recoverable mistakes into systematic failures \citep{zhang2023languagemodelhallucinationssnowball}. Hallucinations in LLMs are hard to avoid given current architectures \citep{xu2025hallucinationinevitableinnatelimitation}, and even the most capable closed-source models struggle to sustain coherent plans over extended horizons \citep{gonzalezpumariega2025robotouilleasynchronousplanningbenchmark}. Thus, effective long-horizon reasoning demands not only accurate tool use and instruction following capabilities, but also a proactive, strategic mindset to adaptively reflect and replan  for more reliable outputs \citep{erdogan2025planandact}.

 To address the challenges, \textbf{Single-agent systems (SAS)}, pioneered by the ReAct paradigm \citep{yao2023react}, operate in a think–act–observe loop until a final answer is produced. Their strength lies in fluid, end-to-end control, where a single model manages reasoning, planning, and tool use, often augmented by post-trained, tool-integrated reasoning models \citep{comanici2025gemini25, li2025chainofagentsendtoendagentfoundation} or inference-time "thinking" modules \citep{anthropic2025thinktool}. However, SAS are limited by a unified context window—as trajectories lengthen, evidence may be truncated or misweighted, leading to premature conclusions and superficial self-reflection \citep{DBLP:conf/icml/DingZZXSX0Y24, chakraborty2025healempiricalstudyhallucinations, hong2025context, liu2025oatzero}.

In contrast, \textbf{multi-agent systems (MAS)} distribute tasks among specialized agents through decentralized handoffs or hierarchical coordination \citep{liang-etal-2024-encouraging, wu2024autogen, tran2025multiagentcollaborationmechanismssurvey}. These systems achieve state-of-the-art results on challenging agentic benchmarks \citep{huang2025deepresearchagentssystematic, wei2025browsecompsimplechallengingbenchmark}, benefiting from explicit separation of context and roles—dedicated agents for planning, reflection, and execution that collectively enhance strategic reasoning \citep{tran2025multiagentcollaborationmechanismssurvey}. In practice, \textbf{human-in-the-loop (HITL)} interventions are often integrated into MAS \citep{11121706}, where specific workflows are designed to allow human operators to pause execution, alter context, or inject strategic signals (Figure~\ref{fig:agent_demo}) \citep{li2025llmbasedautomatedgradinghumanintheloop}. While effective, these interventions rely on direct human feedback and manual design of when and how to inject signals, which are inherently unscalable and unreliable to align with the agent's evolving reasoning context.

While SAS are generally preferred for their fully autonomous nature—enabling greater extensibility and generalizability—existing context management techniques within SAS \citep{DBLP:conf/icml/DingZZXSX0Y24, hong2025context} still lack the explicit role and context separation that MAS architectures provide. To preserve the architectural advantages of MAS—dedicated strategic reasoning agents with isolated context—while maintaining the simplicity and autonomy of SAS, we introduce \textbf{COMPASS} (\emph{Context-Organized Multi-Agent Planning and Strategy System}), a hierarchical framework that decouples tactical execution from strategic oversight through three collaborative components: a \emph{Main Agent} that performs ReAct-style reasoning under dynamically refreshed context, a \emph{Meta-Thinker} that asynchronously monitors the main agent’s progress and issues strategic interventions, and a \emph{Context Manager} that compresses full histories among turns into concise, structured briefs to maintain organized context flow across reasoning stages for stable long-horizon performance.

We evaluate \textsc{COMPASS} on challenging academic and deep-research benchmarks, introduce a test-time scaling extension (\textsc{COMPASS-TTS}), and present the \textsc{Context-12B} model for efficient context management. In sum, our contributions are: (1) \textbf{Formalizing strategic reasoning and context for LHT.} We formalize long-horizon strategic reasoning and highlight its dependence on explicit context management. (2) \textbf{The COMPASS framework.} We propose a hierarchical architecture that separates tactical execution from strategic oversight through explicit role and context separation, enabling reliable operation in error-prone long-horizon settings. (3) \textbf{Comprehensive evaluation and extensions.} We demonstrate \textsc{COMPASS}'s effectiveness across benchmarks, with analyses and practical guidelines for scalable LLM agents.

\section{Agents in Long-Horizon Tasks}
\label{sec:framework}

\textbf{LLM agents.} We define an \textbf{LLM agent} as an autonomous system that leverages a large language model to iteratively reason, act, and observe in service of a goal. At each step $t$, the agent's behavior is conditioned on a context $C_t$, which comprises two components: a \textbf{static context} $C^{\text{static}}$, containing fixed information such as the initial query and tool specifications, and a \textbf{dynamic context} $C^{\text{dyn}}_t$, which accumulates execution traces including thoughts, tool calls, and observations. A single-agent system evolves this context through a thought–action–observation loop:
\[
\textsc{SingleAgent}: C_0 \xrightarrow{(r_0,a_0,o_0)} C_1 \xrightarrow{\cdots} \text{Answer},
\]
while a \emph{multi-agent system (MAS)} distributes the task across several coordinated agents, each with its own context $C^{(i)}_t$, orchestrated through mechanisms such as hierarchical control or peer-to-peer handoffs:
\[
\textsc{MultiAgent}: \{C^{(1)}_t,\ldots,C^{(n)}_t\} \xrightarrow{\;\mathcal{F}\;} \text{Answer}.
\]

\begin{table}[t]
\begin{flushright}
\begin{minipage}{1\linewidth}
\centering
\footnotesize
\setlength{\tabcolsep}{3pt}
\renewcommand{\arraystretch}{1.15}
\caption{\textbf{Illustrative outcomes of meta-thinking decisions.} Each block shows correct and incorrect choices under two common scenarios of long horizon reasoning. Specific case studies are presented in Appendix \S\ref{app:case_all}.}
\label{tab:meta-thinking}
\begin{tabular}{p{0.28\linewidth}|p{0.32\linewidth}|p{0.32\linewidth}}
\hline
\multicolumn{3}{c}{\textbf{Scenario 1: Handling Execution Failures}} \\
\hline
\textbf{Ground Truth $\backslash$ Decision} & \textbf{Continue} & \textbf{Revise} \\
\hline
\textbf{Local Error} & 
\cellcolor{green!15} Correct (Persist) & 
\cellcolor{red!15} Incorrect (Unnecessary Revision) \\
\textit{Example: Search query too narrow} & 
\textit{Refine terms} & 
\textit{Abandon entire approach} \\
\hline
\textbf{Global Dead-End} & 
\cellcolor{red!15} Incorrect (Persistence) & 
\cellcolor{green!15} Correct (Revision) \\
\textit{Example: API permanently broken} & 
\textit{Retry broken API} & 
\textit{Switch to new source} \\
\hline
\multicolumn{3}{c}{} \\[-6pt]
\hline
\multicolumn{3}{c}{\textbf{Scenario 2: Deciding Completion}} \\
\hline
\textbf{Ground Truth $\backslash$ Decision} & \textbf{Conclude} & \textbf{Continue} \\
\hline
\textbf{Correct Solution} & 
\cellcolor{green!15} Correct (Conclude) & 
\cellcolor{red!15} Incorrect (Overthinking) \\
\textit{Example: Verified optimal solution} & 
\textit{Return result} & 
\textit{Keep exploring} \\
\hline
\textbf{Incorrect Solution} & 
\cellcolor{red!15} Incorrect (Premature Stop) & 
\cellcolor{green!15} Correct (Recovery) \\
\textit{Example: Sign error in calculation} & 
\textit{Submit wrong answer} & 
\textit{Re-check computation} \\
\hline
\end{tabular}
\end{minipage}
\end{flushright}
\end{table}

\noindent \textbf{Long-horizon tasks.} We define a task as \textbf{long-horizon} if its successful completion requires a substantial sequence of interdependent reasoning and action steps (e.g., $>10$), often involving iterative tool use, synthesis of intermediate results, and dynamic revision of plans. The principal challenge lies in managing the dynamic context: as the execution trace grows—often linearly with time, $|C^{\text{dyn}}_t| \propto O(t)$—it can exceed the model’s finite context window, obscuring earlier but potentially essential information.

\noindent \textbf{Plans.} Long-horizon tasks typically begin with an initial \emph{plan}, either implicitly generated by the agent or explicitly provided in the context \citep{huang2024understandingplanningllmagents}. Formally, a plan is a sequence of steps $(s_1, s_2, \ldots, s_T)$, where each $s_i$ specifies a concrete action or guideline. Such plans provide a useful backbone but are rarely sufficient on their own \citep{sun2023adaplanner}: unexpected tool responses or initial oversights often require adaptation.

\noindent \textbf{Tactical reasoning.} Given the current step $s_i$ from planning and the dynamic context $C_t$, tactical reasoning determines how to execute $s_i$ to produce useful outputs such as reasoning traces $r_t$ or tool responses $o_t$:
\[
(r_t, o_t) = f_{\text{tac}}(s_i, C_t).
\]
This process assumes the current plan remains valid, and focuses on accurate instruction following and step-level tool execution to achieve the tasks.

\noindent \textbf{Strategic reasoning.} Conditioned on the evolving context $C_t$, strategic reasoning monitors past reasoning for anomalies or inconsistencies and determines whether adjustments are required for subsequent planning:
\[
(s_1, \ldots, s_T)' = f_{\text{strat}}(s_{1:i}, C_t).
\]
If no anomaly is detected, execution continues as planned; if issues are identified, the next trajectory is revised to incorporate corrections or new information, such as an additional verification or backtracking to a earlier stage. If the reasoning already supports a sufficient solution, the process terminates with a final \texttt{<answer>}. Therefore, Long-horizon performance  depends on the interplay between \emph{tactical precision} at each step and \emph{strategic oversight} across steps. Tactical reasoning ensures faithful local execution, while strategic reasoning governs when to correct, adapt, or conclude. Table~\ref{tab:meta-thinking} illustrates some confusion cases, and full examples are covered in in Appendix \S \ref{app:case_all}.

\section{Methods}
\label{sec:compass}

\begin{algorithm*}[t]
\caption{COMPASS: Dual-Loop with Meta Oversight and Context Management.}
\label{alg:compass}
\begin{algorithmic}[1]
\Require Query $q$, tools $\mathcal{O}$, agents $\mathcal{A}^{\text{main}}, \mathcal{A}^{\text{meta}}, \mathcal{A}^{\text{ctx}}$, max iterations $T_{\max}, I_{\max}$
\Ensure Solution $y^*$ for query $q$
\State $n_0 \gets \mathcal{A}^{\text{meta}}.\text{Initialize}(q)$ \Comment{Initial knowledge and planning as \emph{notes}}
\For{$t = 0, 1, \ldots, T_{\max}-1$} \Comment{Outer loop of strategic reasoning and context update}
  \State $x_t \gets \mathcal{A}^{\text{ctx}}.\text{SynthesizeContext}(n_t, q)$
  \State $\mathcal{T}_t \gets \text{ExecuteTurn}(\mathcal{A}^{\text{main}}, \mathcal{A}^{\text{meta}}, x_t, I_{\max})$ \Comment{ReAct loop with monitoring; see Alg.~\ref{alg:inner}}
  \State $\textit{decision} \gets \mathcal{A}^{\text{meta}}.\text{GetDecision}(\mathcal{T}_t,x_t)$ \Comment{Strategic decision from meta-thinking}
  \If{$\textit{decision}=\textsc{stop}$ \textbf{or} $t = T_{\max}-1$}
    \State \Return $\text{ExtractAnswer}(\mathcal{T}_t, n_t)$
  \EndIf
  \State $n_{t+1} \gets \text{UpdateNotes}(n_t, \mathcal{T}_t, \textit{decision})$
\EndFor
\end{algorithmic}

\vspace{2pt}
\noindent\footnotesize 
\textbf{ExtractAnswer}: re-summarizes the final trace $\mathcal{T}_t$ with context from notes $n_t$ to produce the \texttt{<answer>} (see App.\S~\ref{app:answersynth}).\\
\textbf{UpdateNotes}: appends the current context in a structured form to the rolling set of prior briefs $\{x_0,\ldots,x_{t-1}\}$ (see App.\S~\ref{app:notestore}).
\end{algorithm*}

\subsection{Architectural Overview}
\label{subsec:architecture_overview}

At the core of \textsc{COMPASS} are three specialized agents with clearly separated responsibilities:

\noindent \textbf{Main Agent} serves as the primary executor for \emph{tactical reasoning} with a ReAct-style workflow. At step $t$, it alternates between generating an intermediate thought and producing a tool command appended to the running trace and iteratively continues until the desired answer is obtained.
Viewed in isolation, this resembles the single-agent system defined in Section~\ref{sec:framework}. Within \textsc{COMPASS}, however, the Main Agent is supplied with a \emph{renewed context} to follow from the Context Manager whenever a strategic intervention occurs and can be \emph{interrupted} at any moment by the meta-thinking agent. 

\noindent \textbf{Meta-Thinker} \emph{reasons over previous reasoning}, monitoring the reasoning trajectory together with its dedicated context. It remains silent until anomalies are detected—such as looping behavior, tool misuse, or signs of task completion. Once triggered, it issues a high-level strategic signal to prompt reflection, termination, or verification, which is then passed to the Context Manager. Running asynchronously ensures that these interventions do not block the Main Agent’s reasoning, thereby preserving execution fluidity. Because it operates on a single turn rather than full traces, the Meta-Thinker is designed to be lightweight, capable of catching up with the Main Agent’s operation with low latency, especially when combined with prompt-caching techniques \citep{gim2024promptcachemodularattention}.

\noindent \textbf{Context Manager} serves as the system’s adaptive context controller, responsible for determining what information should enter the active context for Main and Meta-thinking agent at each iteration.
It \emph{synthesizes a new, task-specific context} by selectively drawing from three sources: (i) the persistent \emph{structured notes} accumulated across turns, (ii) the current reasoning trajectory from Main Agent, and (iii) the Meta-Thinker’s strategic signal.
Through this process, the Context Manager provides the Main Agent with a concise and relevant context that preserves continuity while avoiding redundancy or distraction.
After each iteration, the resulting refreshed context is also appended to the evolving note store for future retrieval, ensuring long-horizon coherence with minimal memory overhead.

\begin{figure}[t]
\centering
\vspace{-5pt}
\includegraphics[width=0.98\linewidth]{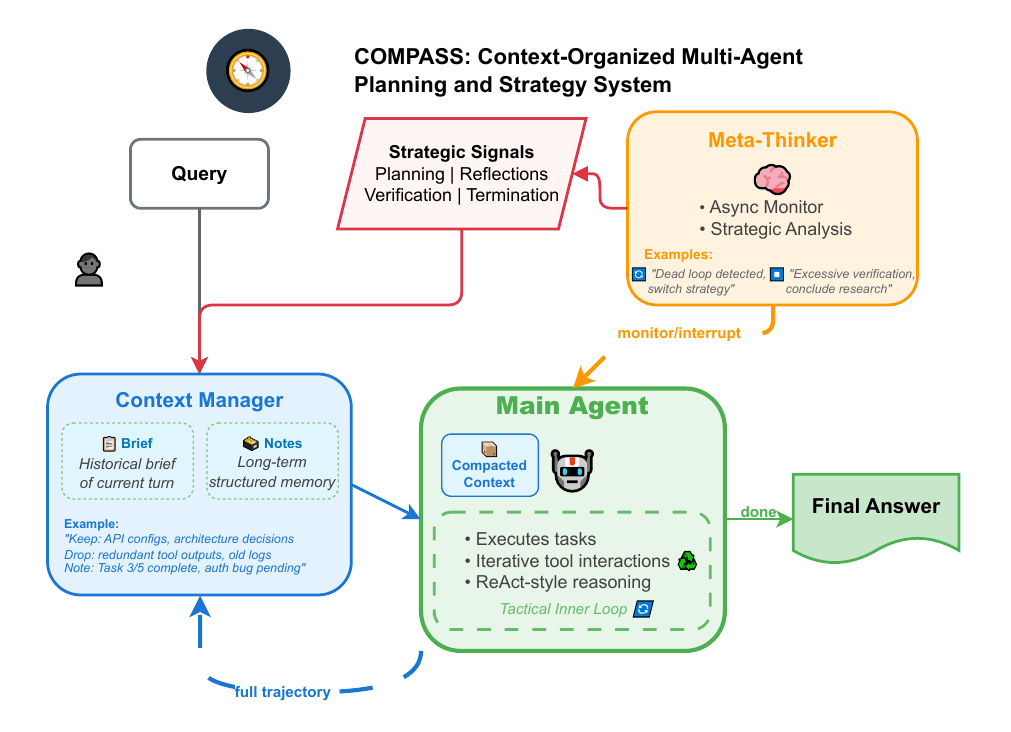}
\caption{\textbf{The COMPASS dual-loop framework.} The \emph{Main Agent} performs tool interactions by following the instructions from continually refreshed context; the \emph{Meta-Thinker} asynchronously monitors trajectories and triggers strategic decisions, and the \emph{Context Manager} compresses full histories (from structured notes) into concise, contextual aware briefs back to Main Agent.}
\label{fig:workflow}
\vspace{-8pt}
\end{figure}

\subsection{Trajectory Lifecycle}

The \textsc{COMPASS} framework (Algorithm~\ref{alg:compass}) begins when the user issues a query $q$, which initializes the working state $h_0 = (q)$ via the meta-thinker. 
Before execution, the Meta-Thinker performs an \emph{initial planning step}, outlining a coarse action sequence or information sources to explore. 
This plan defines the first context $x_0$, serving as a reference for subsequent meta-decisions. 
Execution then proceeds through iterative outer loops of reasoning and oversight, each consisting of two coupled stages:

\noindent \textbf{(1) Reasoning, Action, Monitoring, and Reflection.}  
During each iteration $t$, the Main Agent performs \emph{tactical reasoning and action} using the current context $x_t$, producing intermediate thoughts and tool invocations that expand the trajectory $\mathcal{T}_t$. 
Concurrently, the Meta-Thinker engages in \emph{strategic monitoring}, asynchronously inspecting $\mathcal{T}_t$ for anomalies (e.g., loops, tool misuse, reasoning drift) or completion signals. 
Upon detection, it issues a high-level strategic signal (e.g. replan or verify)  that guides the next context synthesis.

\noindent \textbf{(2) Context Refresh and Note Integration.}  
After a meta interruption, the Context Manager synthesizes a refreshed context 
$x_{t+1} = \text{SynthesizeContext}(q, \mathcal{T}_t, n_t, \textit{decision}_t)$ 
by integrating verified evidence, relevant notes (raw query $q$ and initial plan are included if no dynamic context yet available), and the Meta-Thinker’s signal $\textit{decision}_t$.  
It then appends extracted segments—key observations of what has worked, what has failed, and remaining uncertainties—to the note store (App.\S~\ref{app:notestore}), updating $n_{t+1} = n_t \cup \text{ExtractSections}(x_t)$ for future iterations.

If $\textit{decision}_t = \textsc{stop}$, the Context Manager invokes the Answer Synthesizer (App.\S~\ref{app:answersynth}) to generate the final output; 
otherwise, $(x_{t+1}, n_{t+1})$ are passed forward to initialize the next iteration. This loop continues until convergence or the maximum round $T_{\max}$ is reached, ensuring that long-horizon reasoning proceeds with bounded context size and persistent continuity across turns.

\section{Experimental Results}

\label{sec:experiments}

\paragraph{Benchmarks and Baselines.}
Our evaluation for COMPASS focuses on \textbf{DeepResearch-style long-horizon benchmarks}, typically demanding 20+ reasoning–action steps: (i) \textbf{GAIA} \citep{mialon2024gaia}, including all Level 1–3 non-image tasks; (ii) \textbf{BrowseComp} \citep{wei2025browsecompsimplechallengingbenchmark}, with 1,266 web navigation tasks requiring verification of entangled facts; and (iii) \textbf{Humans' Last Exam (HLE)} \citep{phan2025hlexam}, yielding 2,158 questions across mathematics, humanities, and natural sciences after excluding image-based items. We compare COMPASS against two baseline groups: fundamental paradigms including single-agent systems (Search/Browse tools, +thinking Tool, +Context management tool), multi-agent systems (Manager hierarchical delegation and Decentralized Handoffs coordination with same set of agent as a tool), and Iterative Refinement workflows \citep{wang-etal-2023-plan}); Established research agents (OpenAI's DeepResearch, DeepSeek's Agent V3.1, Google's TestTime Diffusion) are included to compare with the Test-time scaling option (See Section \ref{subsec:parallel-sampling}). All experiments use \textbf{Gemini 2.5 Pro/Flash} and as backbone reasoning models and native Google serach/browsing and code execution tool is used.

\paragraph{Evaluation Metrics.}
Our primary metric is \textbf{Pass@1 accuracy}. We also assess strategic reasoning through four trajectory-level metrics, motivated by the failure mode in Table \ref{tab:meta-thinking}: \textbf{Persist Appropriateness Rate (PAR)} measures whether agents appropriately continue with valid plans, \textbf{Pivot Recognition (PVR)} captures whether agents pivot when current approaches fail, \textbf{Conclude Accuracy (CA)} indicates whether agents correctly recognize when to halt with a solution, and \textbf{Error-Recovery Continuation (ERC)} reflects whether agents continue searching after incorrect answers. These metrics, evaluated via LLM-as-a-Judge with structured outputs (App.\S~\ref{app:prompts}), expose precision-recall trade-offs in long-horizon problem solving—high PAR without PVR indicates blind adherence to failing plans, while high CA without ERC leads to premature termination—revealing more fine-grained details on how agents navigate critical decision points.



\begin{table}[t!]
\centering
{\fontsize{7.6pt}{8.8pt}\selectfont
\setlength{\tabcolsep}{2.8pt}
\renewcommand{\arraystretch}{0.9}

\definecolor{HighlightGray}{gray}{0.92}
\newcommand{\rothead}[1]{\rotatebox{90}{\strut #1}}
\newcommand{\goo}[1]{\raisebox{-0.1ex}{#1}}
\newcommand{\srch}{\textcolor{blue!70!black}{\goo{\faSearch}}}
\newcommand{\code}{\textcolor{green!60!black}{\goo{\faCode}}}
\newcommand{\term}{\textcolor{orange!80!black}{\goo{\faTerminal}}}
\newcommand{\brws}{\textcolor{purple!70!black}{\goo{\faGlobe}}}
\newcommand{\thnk}{\textcolor{red!70!black}{\goo{\faBrain}}}
\newcommand{\cntx}{\textcolor{teal!70!black}{\goo{\faDatabase}}}

\newcommand{\deepseeklogo}{\raisebox{-0.45ex}{\includegraphics[height=2.5ex]{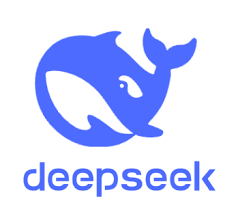}}}
\newcommand{\openailogo}{\raisebox{-0.35ex}{\includegraphics[height=2.1ex]{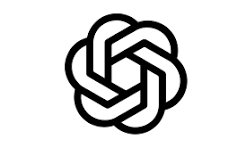}}}
\newcommand{\googlelogo}{\raisebox{-0.40ex}{\includegraphics[height=1.8ex]{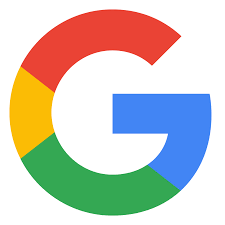}}}

\sisetup{
  detect-weight,
  mode=text,
  group-digits=false,
  table-number-alignment=center
}
\newcolumntype{Y}{S[table-format=2.1, table-column-width=7.6mm]}
\newcolumntype{P}[1]{>{\raggedright\arraybackslash}p{#1}}

\captionsetup{font=small}
\caption{Pass@1(\%) are reported. Datasets:\textbf{BC}=\textit{BrowseComp}, \textbf{GAIA}, \textbf{HLE}. Tools: 
\textcolor{blue!70!black}{\faSearch} Search,\;
\textcolor{green!60!black}{\faCode} Coding,\;
\textcolor{orange!80!black}{\faTerminal} Terminal,\;
\textcolor{purple!70!black}{\faGlobe} Browsing,\;
\textcolor{red!70!black}{\faBrain} Thinking (for thinking over previous reasoning),\;
\textcolor{teal!70!black}{\faDatabase} Context (for keeping the most relevant context).}
\label{tab:performance_single}

\begin{tabular}{@{} P{2.6cm} c Y Y Y @{}}
\toprule
\textbf{Method} & \textbf{Tools} &
\rothead{\textbf{BC}} & \rothead{\textbf{GAIA}} & \rothead{\textbf{HLE}} \\
\midrule

\multicolumn{5}{@{}l}{\textbf{Gemini 2.5 Pro}}\\[-0.4ex]
\multicolumn{5}{@{}l}{\textsc{Single-Agent}}\\
Search and Browse Only              & \srch\,\brws                         & 16.8 & 58.6 & 14.8 \\
+ Meta-Thinking          & \srch\,\brws\,\thnk                  & 26.4 & 61.6 & 20.6 \\
+ Context                & \srch\,\brws\,\thnk\,\cntx           & 29.8 & 63.1 & 28.4 \\
\cmidrule(l){1-5}
\multicolumn{5}{@{}l}{\textsc{Multi-Agent}}\\
Agent-as-a-Tool          & \srch\,\brws\,\code\,\thnk\,\cntx    & 31.8 & 65.5 & 28.6 \\
Decentralized Handoffs   & \srch\,\brws\,\code\,\thnk\,\cntx    & 28.1 & 64.8 & 28.3 \\
Iterative Refinement     & \srch\,\brws\,\code                  & 30.5 & 65.3 & 27.9 \\
\rowcolor{HighlightGray}
\textbf{COMPASS}         & \srch\,\brws\,\code\,\thnk\,\cntx    & \bfseries 35.4 & \bfseries 67.8 & \bfseries 31.7 \\
\midrule

\multicolumn{5}{@{}l}{\textbf{Gemini 2.5 Flash}}\\[-0.4ex]
\multicolumn{5}{@{}l}{\textsc{Single-Agent}}\\
Search and Browse Only              & \srch\,\brws                         & 12.1 & 53.5 & 12.2 \\
+ Meta-Thinking          & \srch\,\brws\,\thnk                  & 19.8 & 56.3 & 18.7 \\
+ Context                & \srch\,\brws\,\thnk\,\cntx           & 22.6 & 58.6 & 22.9 \\
\cmidrule(l){1-5}
\multicolumn{5}{@{}l}{\textsc{Multi-Agent}}\\
Agent-as-a-Tool          & \srch\,\brws\,\code\,\thnk\,\cntx    & 23.4 & 58.9 & 22.1 \\
Decentralized Handoffs   & \srch\,\brws\,\code\,\thnk\,\cntx    & 21.9 & 58.1 & 23.8 \\
Iterative Refinement     & \srch\,\brws\,\code                  & 22.8 & 58.7 & 23.3 \\
\rowcolor{HighlightGray}
\textbf{COMPASS}         & \srch\,\brws\,\code\,\thnk\,\cntx    & \bfseries 26.1 & \bfseries 60.2 & \bfseries 24.6 \\
\midrule

\multicolumn{5}{@{}l}{\textit{\textbf{Established Research Agents}}}\\
DeepResearch (o3)  & \srch\,\brws\,\code\,\term & \bfseries 51.1 & 67.4 & 26.6 \\
\cmidrule{1-5}
DeepSeek V3.1 Terminus Agent & \srch\,\brws\,\code\,\term  & 38.5 & 63.1 & 21.7 \\
\cmidrule{1-5}
Test Time Diffusion (Gemini 2.5 Pro) & \srch\,\brws\,\code & {---} & 69.1 & 33.9 \\
\cmidrule{1-5}
\rowcolor{HighlightGray}
\textbf{COMPASS-TTS} (Gemini 2.5 Pro) & \srch\,\brws\,\code\,\thnk\,\cntx & 43.7 & \bfseries 72.1 & \bfseries 35.2 \\
\bottomrule
\end{tabular}
}
\end{table}

\subsection{Main Results.} Our main results (Table~\ref{tab:performance_single}) reveal a clear path to enhancing agent capabilities. We first establish the importance of \textbf{structured reasoning and context}, as augmenting a single agent with meta-thinking and context management capabilities consistently improves its performance. Building on this, our \textbf{COMPASS} architecture further amplifies these gains by externalizing these roles into \textbf{dedicated agents for explicit monitoring, thinking, and context curation}---a method that shown more effectiveness than taking these functions as tools in single agent. Finally, these architectural benefits deliver \textbf{consistent performance} improvements across different models and benchmarks, with gains most pronounced on BrowseComp, where tasks require sustained multi-source navigation in long horizon, achieving results that are comparable to well-established deep-research agents with our framework when scaled with sampling.

\begin{figure}
    \centering
    \includegraphics[width=\linewidth]{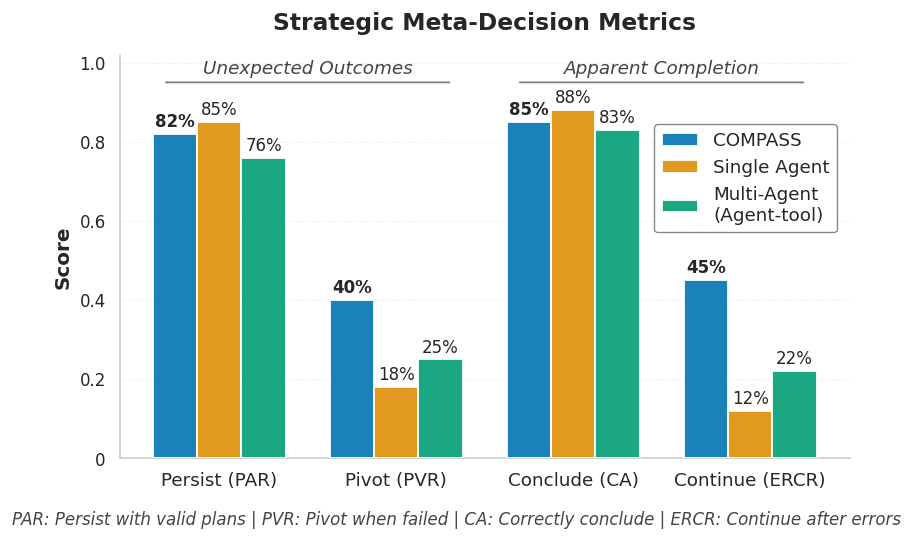}
    \caption{
        \textbf{Strategic meta-decision metrics across agent variants.} 
        Bars report scores for four metrics to measure streatgic reasoning: 
        Persist (PAR) and Pivot (PVR), and 
        Conclude (CA) and Continue (ERCR); see Table \ref{tab:meta-thinking} for examples and formal definitions.}
    \vspace{-5pt}
    \label{fig:agent_performance}    \vspace{-10pt} 
\end{figure}

\noindent \textbf{Strategic Reasoning and Case Analysis.} Figure~\ref{fig:agent_performance} 
reveals how architectural separation improves strategic behavior. 
Single-agent baselines exhibit high PAR but low PVR—\emph{blind 
persistence} where constraints become buried under accumulated outputs. 
Multi-agent baselines improve pivoting but sacrifice persistence, 
suggesting coordination alone is insufficient without context curation.

COMPASS balances these through complementary mechanisms: the Meta-Thinker 
detects anomalies (looping, tool misuse) before error cascade, while 
the Context Manager maintains concise briefs that preserve constraints. 
\textbf{Case studies (Appendix~\S\ref{app:case_all}) demonstrate this 
prevents two failure modes common in Re-Act style agentic system—context overload causing premature conclusions 
(\S\ref{app:incorrect_continue}) and contextual pollution causing repeated 
errors (\S\ref{app:local_error_persist})—while enabling appropriate 
persistence during refinement (\S\ref{app:local_error_persist}) and 
strategic pivoting at dead ends (\S\ref{app:global_deadend_revise}).}

\paragraph{Ablation Studies.}
Table~\ref{tab:combined_ablation_browsecomp} systematically ablates Meta-Thinking and Context Management components on BrowseComp, revealing distinct failure modes and scaling behaviors. Removing the Meta-Thinker entirely collapses adaptive capability, causing \emph{blind persistence} with high task completion but near-zero strategic metrics. Scaling up the Meta-Thinker shows systematic improvements in performance and strategic reasoning while maintaining token efficiency, indicating that \emph{oversight quality} matters more than raw capacity. Context Manager ablations reveal different trade-offs: removing it causes token bloat from repeatedly revisiting failed attempts, while certain configurations like Gemini 2.5 Flash exhibit a failure mode of \emph{excessive plan revision}, triggering strategic interventions that extend execution while improving adaptability. This highlights that context curation requires balancing compression with strategic signaling rather than pure summarization. Neither component alone achieves the full system's balanced performance across strategic metrics, confirming that oversight and context management are \emph{complementary capabilities} that work synergistically for effective long-horizon reasoning.

\begin{table*}[t]
\centering
\caption{Ablation studies for the Meta-Thinking and Context Management components on \textbf{BrowseComp}. The full system is presented as a reference. Each subsequent section ablates one component, showing a marked drop in success and strategic metrics compared to the full system. \textbf{Strategy Adequacy} represents the average of 4 metrics.}
\label{tab:combined_ablation_browsecomp}
\resizebox{\linewidth}{!}{
\begin{tabular}{lcccccccr}
\toprule
\textbf{Component Configuration} & \textbf{Pass @ 1 (\%)} & \textbf{Component Tokens*} & \textbf{Total Tokens} & \textbf{PAR} & \textbf{PVR} & \textbf{CA} & \textbf{ERC} & \textbf{Strategy Adequacy} \\
\midrule
\multicolumn{9}{l}{\textit{\textbf{Reference: Full System}}} \\
Gemini 2.5 Pro (Full System) & \textbf{35.4} & 59K & 185K & 0.85 & \textbf{0.48} & 0.88 & \textbf{0.55} & \textbf{0.69} \\
\midrule
\multicolumn{9}{l}{\textit{\textbf{Ablation 1: Meta-Thinking Agent} (Context Manager/Main Agent fixed to Gemini 2.5 Pro)}} \\
None (Main + Context only) & 15.2 & 0K & 85K & \textbf{0.92} & 0.12 & \textbf{0.95} & 0.21 & 0.55 \\
Gemini 2.5 Flash & 28.5 & 8K & 132K & 0.79 & 0.33 & 0.78 & 0.38 & 0.57 \\
Gemini 2.5 Pro w/ Extra Tools & 32.8 & 14K & 140K & 0.82 & 0.41 & 0.76 & 0.45 & 0.61 \\
\midrule
\multicolumn{9}{l}{\textit{\textbf{Ablation 2: Context Manager} (Meta-Thinking/Main Agent fixed to Gemini 2.5 Pro)}} \\
None (No Context Manager) & 26.4 & 0K & 156K & 0.91 & 0.18 & 0.94 & 0.25 & 0.57 \\
Gemma-3-12B & 28.5 & 12K & 144K & 0.77 & 0.35 & 0.76 & 0.44 & 0.58 \\
Gemini 2.5 Flash & 31.8 & 20K & 212K & 0.74 & 0.52 & 0.80 & 0.54 & 0.65 \\
\bottomrule
\multicolumn{9}{l}{\footnotesize{*Component Tokens measure usage for the specific agent being ablated (Meta-Thinking or Context Manager). For the Total, it's the sum of all three agents.}}
\end{tabular}
}
\end{table*}

\section{Practical Extensions}

In addition to the core framework, we introduce two extensions that further enhance robustness and efficiency: (1)  a specialized compact context manager, \textbf{Context-12B}, trained via supervised fine-tuning (SFT) and direct preference optimization (DPO) to reduce token cost while maintaining strong performance , and (2) a test-time scaling variant, \textbf{COMPASS-TTS}, which leverages parallel sampling to improve reliability under uncertainty.

\subsection{Context-12B: Training Specialized Context Managers}
\label{sec:training}

While larger models excel at summarization and context organization, their high API costs and deployment overhead hurts the efficiency of the system. Our case analyses as shown in Appendix \S\ref{app:case_study_comparison} further revealed that, among the three COMPASS agents, the \emph{Context Manager} is the most structured and deterministic, operating more like a summarizer than an open-ended reasoner. This observation motivates us to design a smaller, deployable model that retains the research-status-oriented summarization ability of larger models without their computational footprint.

\noindent \textbf{Data Collection.}
We leverage \textbf{Gemini 2.5 Pro} with searching tool as a \emph{data engine} to generate high-quality training signals. Using COMPASS rollouts on Knowledge-intense long-horizon benchmarks or complex academic QA (GAIA, SimpleQA, MMLU-Pro, etc), we extract training data where the input is the full reasoning trajectory plus meta-thinking notes, and the output is the optimized brief provided to the Main Agent. To ensure quality, we apply filtering:
(i) remove trivial trajectories with fewer than three tool interactions,
(ii) exclude degenerate completions where the correct answer is reached without reflection,
and (iii) upsample cases where context management clearly drives recovery or proper task termination.

\noindent \textbf{Training Pipeline.}
We first distill this capability into \textbf{Gemma-3 12B}, obtaining \textbf{Context-12B-SFT}.
Supervised fine-tuning teaches the model to follow the instructions and produce concise, strategically aligned briefs from complex trajectories from bigger models with our collected data. We then refine Context-12B-SFT using \textbf{direct preference optimization (DPO)}. For each training trajectory, we sample multiple candidate summaries and construct preference pairs from the same data we used for SFT: We continue applying the generated contexts in the COMPASS inference engine, and selecting the context leading successful completions with fewer tokens are labeled as preferred, while redundant or error-prone summaries are rejected.(more training details can be referred in Appendix \S\ref{app:training}) This produces the final model, \textbf{Context-12B}, optimized for both accuracy and efficiency.

\begin{figure}
     \centering 
    \includegraphics[width=0.48\textwidth]{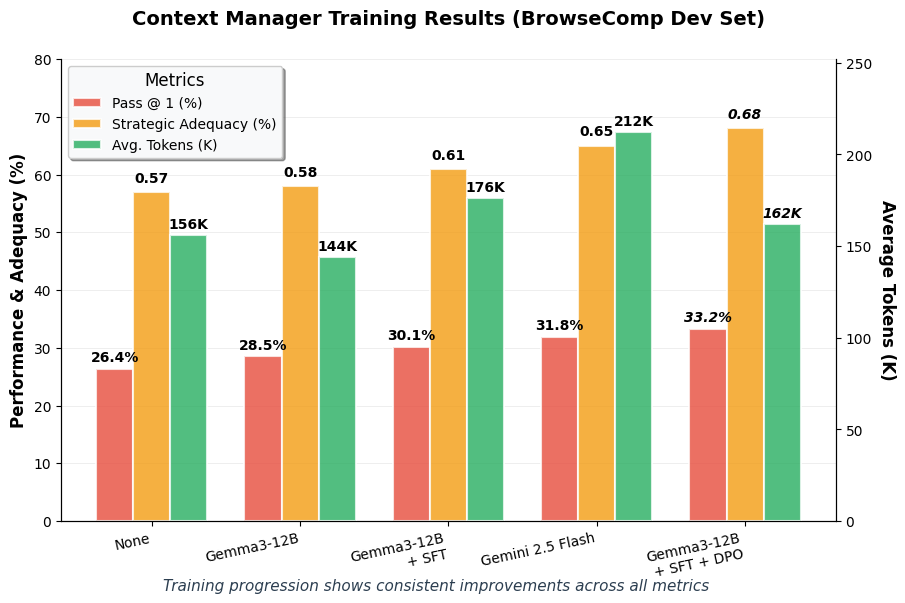} 
    \caption{Context-12B Performance on BrowseComp. Pass @ 1(\%), strateg adequacy (×100), and token efficiency all improve progressively. DPO yields substantial efficiency gains without sacrificing accuracy.}
    \label{fig:context_manager_training}
    \vspace{-12pt} 
\end{figure}

\paragraph{Results.}
As shown in Figure~\ref{fig:context_manager_training}, we evaluated the performance of Context-12B on Browsecomp, achieving performance comparable to a larger models (Gemini 2.5 Flash) while using only 70\% of their tokens with a SFT-DPO training pipeline .

\begin{figure*}[t]
\centering
\includegraphics[width=0.90\linewidth]{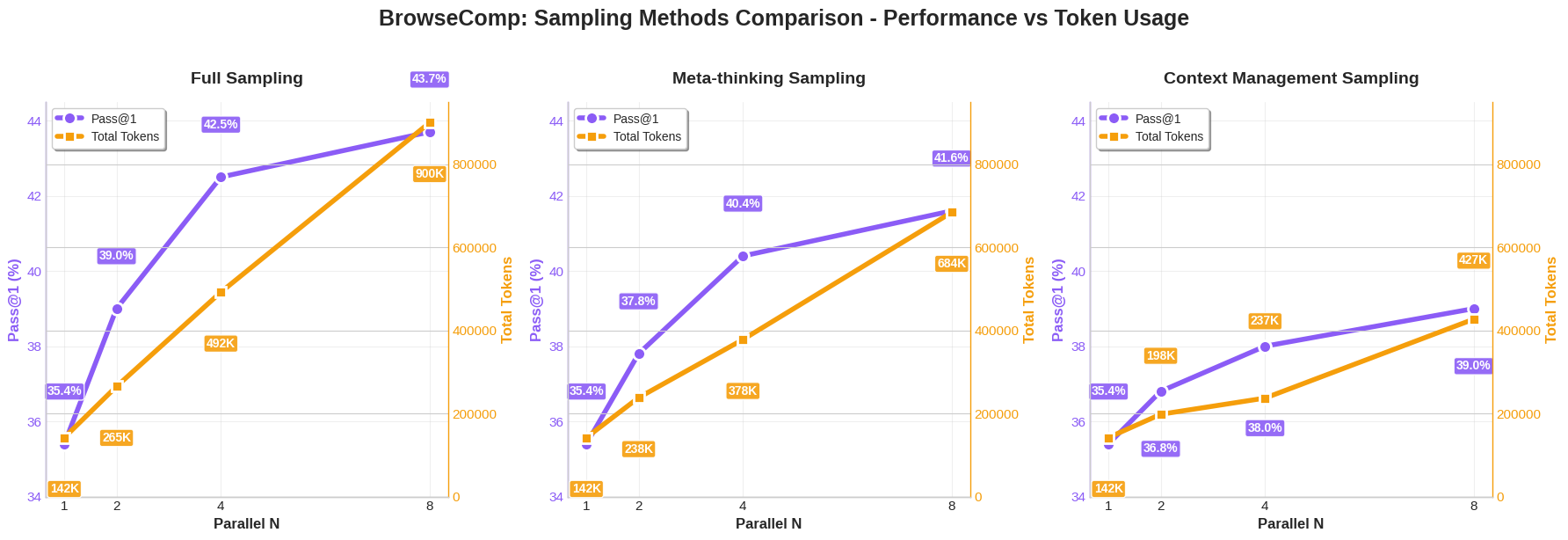}
\caption{Performance (Pass@1) vs. token cost for three COMPASS-TTS sampling methods on the BrowseComp benchmark. Increasing the number of parallel samples improves accuracy but also raises token costs.}
\vspace{-8pt}
\label{fig:parallel}
\end{figure*}

\subsection{COMPASS-TTS: Test-time Scaling with Parallel Sampling}
\label{subsec:parallel-sampling}
Parallel sampling during inference is a common strategy to improve the performance~\citep{wang2023selfconsistency,han2025deep}. With $n$ as parallel samples, we extend the framework with \textbf{COMPASS-TTS}, which explores multiple reasoning or context-management alternatives concurrently as (1) \textbf{Full-pipeline sampling (Full-PS).} Executes $n$ diversified runs of the \emph{entire pipeline} (varying seeds and temp). A lightweight synthesizer $g$ then aggregates the candidate outputs into a single final answer.  (2) \textbf{Meta-thinking sampling (MT-PS).} Parallelizes only the meta-thinking module, producing $n$ alternative triggering and decision proposals. A synthesizer with slightly different prompt than the Full-PS's one merges these into one coherent downstream plan, which is then executed once.  (3) \textbf{Context-management sampling (CM-PS).} Parallelizes only the context manager, yielding $n$ alternative contextualization. These are distilled by the synthesizer into a single injected context before task execution.

\noindent \textbf{Results.} Figure~\ref{fig:parallel} plots accuracy and total tokens for $n \in \{1,2,4,8\}$. All methods improve monotonically over $n=1$, with Full-PS showing the strongest gains but steepest token growth, MT-PS offering a balanced middle ground, and CM-PS achieving the best efficiency. Performance plateaus around $n=4$, suggesting that $n=2$–$4$ offers a practical sweet spot to balance accuracy and costs.

\section{Related Work}
\label{sec:related_work}

\textbf{LLM-based Agentic Systems.}
LLM-based agents primarily follow two paradigms. \textbf{Single-agent systems} extend the ReAct-style think–act–observe loop \citep{yao2023react, shinn2023reflexion} into multi-turn settings. They are popular for their simplicity and autonomy, and have been enhanced via reinforcement learning \citep{zhang2025landscape} to yield post-trained, tool-integrated models \citep{comanici2025gemini25, li2025chainofagentsendtoendagentfoundation, feng2025retoolreinforcementlearningstrategic, jin2025search} and inference-time reasoning and memory modules \citep{anthropic2025thinktool, wang2024voyager}.  
\textbf{Multi-agent systems (MAS)} instead distribute reasoning across specialized roles for improved robustness \citep{liang-etal-2024-encouraging, wu2024autogen, tran2025multiagentcollaborationmechanismssurvey}. MAS employ explicit coordination by centralized managers, decentralized handoffs, or predefined workflows \citep{anthropic2024effectiveagents} and often achieve state-of-the-art results on complex benchmarks \citep{huang2025deepresearchagentssystematic, wei2025browsecompsimplechallengingbenchmark, snell2025scaling, li2023camel}. However, their reliance on manually designed pipelines limits scalability and generalization \citep{Sapkota_2026, pan2025why}. Our work instead makes \emph{strategic reasoning} and \emph{context management} explicit architectural components while retaining the fluid, end-to-end nature of single-agent systems.

\noindent \textbf{LLM Reasoning for Long-Horizon Tasks.}
While Chain-of-Thought (CoT) reasoning improves short reasoning tasks \citep{wei2022chain}, it falters on long-horizon problems where errors accumulate over multiple steps \citep{chen2025reasoningerasurveylong, zhang2023languagemodelhallucinationssnowball}. Recent efforts enhance long-horizon reasoning through hierarchical control \citep{wang2025emergenthierarchicalreasoningllms, chen2025reasoningerasurveylong} and explicit planning or self-reflection, using prompting \citep{sun2023adaplanner, madaan2023selfrefine}, post-training \citep{deepseekr1, parmar2025plantuningposttraininglanguagemodels}, or both \citep{erdogan2025planandact}. Yet, these methods often remain brittle or superficial \citep{liu2025oatzero, lindsey2025biology, huang2024large}, especially under incomplete or excessive context. Insights from \emph{context engineering} studies highlight how redundant inputs can degrade reasoning quality \citep{hong2025context, mei2025surveycontextengineeringlarge, li2025thinkingfailspitfallsreasoning, liu-etal-2024-lost}. Motivated by this, our framework integrates adaptive context selection into the reasoning process to mitigate cognitive load and improve adaptivity under long horizon.

\section{Conclusion}
\label{sec:conclusion}

We presented \textbf{COMPASS}, a hierarchical framework that elevates strategic reasoning and context management to \emph{architectural primitives}, enabling reliable long-horizon reasoning without the complexity of multi-agent topologies. Our findings highlight the critical distinction between \emph{tactical reasoning}—the capacity for accurate tool use and instruction following within an agent turn—and \emph{strategic reasoning}—the higher-level oversight that guides reflection, replanning, and termination decisions. We further highlight the role of \emph{context management} in maintaining coherence and adaptability across extended trajectories, offering actionable principles for developing scalable and robust agentic systems in long-horizon reasoning tasks.

\newpage 

\section*{Limitations}
\label{sec:limitations}

While \textsc{COMPASS} demonstrates strong performance on QA-style agentic benchmarks, our evaluation currently emphasizes controlled reasoning environments to clearly illustrate the framework’s proof of concept. Specifically, experiments are conducted within settings that involve searching, text-based browsing, and code execution tools. Future extensions to more open-ended domains—and the integration of richer interoperability mechanisms such as MCP servers and agent-to-agent (A2A) communication protocols \citep{hou2025modelcontextprotocolmcp}—would enable a more comprehensive assessment of \textsc{COMPASS}'s robustness in dynamic, real-world contexts. In addition, Our study primarily focuses on proprietary frontier models, as our goal is to incentivize and analyze the limits of state-of-the-art reasoning capabilities on challenging long-horizon tasks. Common open-source models currently underperform significantly on such tasks \citep{plaat2025survey}, but future work can extend our framework to investigate their performance and post-training potential.

\bibliography{custom}

\appendix

\section{Reproducibility}

To ensure reproducibility of our results, we provide comprehensive implementation details and experimental specifications throughout this work. Section~\ref{sec:experiments} and Appendix~\ref{app:resources} detail our experimental setup, including specific model versions (Gemini 2.5 Pro/Flash), hyperparameters, and evaluation metrics. All prompt templates for the three COMPASS agents are provided in Appendix~\ref{app:prompts}, along with evaluation prompts for our strategic reasoning metrics. Our training pipeline for Context-12B is fully specified in Appendix~\ref{app:training}, including data preprocessing steps, SFT and DPO hyperparameters, and infrastructure requirements. Detailed ablation studies in Table~\ref{tab:combined_ablation_browsecomp} isolate the contribution of each component. Our evaluation covers three established benchmarks (GAIA, BrowseComp, HLE) with standard Pass@1 metrics, supplemented by novel strategic reasoning metrics that are operationally defined with LLM-as-a-Judge protocols detailed in Appendix~\ref{app:prompts}. Case studies in Appendix~\ref{app:case_all} provide concrete trajectory examples illustrating key failure modes and recovery patterns. All experiments used Google Cloud infrastructure with specific GPU configurations and API endpoints documented in Appendix~\ref{app:resources}. While our implementation relies on proprietary Gemini models, the architectural principles and algorithmic framework are model-agnostic and can be adapted to other LLM backends.

\section{Prompt Templates}
\label{app:prompts}

We report here the full prompt templates used in evaluation and in the definition of 
\textbf{COMPASS}. The evaluation templates are used for accuracy and strategic reasoning 
metrics, while the agent templates specify the architectural roles of COMPASS's components. 

\subsection{Evaluation Prompts}

\paragraph{Accuracy Evaluation.} 
We evaluate correctness (Pass@1) using a query template that enforces a canonical answer format, 
and a grader template that judges against the gold label.

\textbf{Query Template}

\begin{tcolorbox}[colback=gray!5, colframe=black!75, breakable]
\texttt{\{Question\}}

Your response should be in the following format: \\
Explanation: \texttt{\{\{your explanation for your final answer\}\}} \\
Exact Answer: \texttt{\{\{your succinct, final answer\}\}}
\end{tcolorbox}

\textbf{Grader Template}

\begin{tcolorbox}[colback=gray!5, colframe=black!75, breakable]
Judge whether the following [response] to [question] is correct or not 
based on the precise and unambiguous [correct\_answer] below.

\textbf{[question]:} \texttt{\{question\}} \\
\textbf{[response]:} \texttt{\{response\}} \\
\textbf{[correct\_answer]:} \texttt{\{correct\_answer\}}

Your judgement must be in the format and criteria specified below:

\textbf{extracted\_final\_answer:} The final exact answer extracted from the [response]. 
Put 'None' if there is no exact, final answer.

\textbf{reasoning:} Explain why the extracted\_final\_answer is correct or incorrect 
based only on [correct\_answer].

\textbf{correct:} Answer 'yes' if extracted\_final\_answer matches [correct\_answer] 
(with small tolerance for numerical error), otherwise 'no'.
\end{tcolorbox}

\paragraph{Strategic Reasoning Metrics.} 
The following prompts evaluate the four strategic reasoning metrics (PAR, PVR, CA, ERC) 
defined in Section~\ref{sec:framework}. Each template requires both a justification and a 
binary decision. 

\textbf{Persist Appropriateness Rate (PAR)}

\begin{tcolorbox}[colback=gray!5, colframe=black!75, breakable]
\textbf{[state]:} \texttt{\{trajectory\_state\}} \\
\textbf{[decision]:} \texttt{\{decision\}}

\textbf{reasoning:} Explain whether persisting was appropriate in this situation. \\
\textbf{par\_correct:} Answer 'yes' if persisting was the right choice, otherwise 'no'.
\end{tcolorbox}

\textbf{Pivot Recognition (PVR)}

\begin{tcolorbox}[colback=gray!5, colframe=black!75, breakable]
\textbf{[state]:} \texttt{\{trajectory\_state\}} \\
\textbf{[decision]:} \texttt{\{decision\}}

\textbf{reasoning:} Explain whether pivoting was the right action here. \\
\textbf{pvr\_correct:} Answer 'yes' if pivoting was appropriate, otherwise 'no'.
\end{tcolorbox}

\textbf{Conclude Accuracy (CA)}

\begin{tcolorbox}[colback=gray!5, colframe=black!75, breakable]
\textbf{[state]:} \texttt{\{trajectory\_state\}} \\
\textbf{[decision]:} \texttt{\{decision\}} \\
\textbf{[ground\_truth\_answer]:} \texttt{\{correct\_answer\}}

\textbf{reasoning:} Explain whether concluding was accurate given the ground truth. \\
\textbf{ca\_correct:} Answer 'yes' if conclusion matches the correct answer, otherwise 'no'.
\end{tcolorbox}

\textbf{Error-Recovery Continuation (ERC)}

\begin{tcolorbox}[colback=gray!5, colframe=black!75, breakable]
\textbf{[state]:} \texttt{\{trajectory\_state\}} \\
\textbf{[decision]:} \texttt{\{decision\}} \\
\textbf{[ground\_truth\_answer]:} \texttt{\{correct\_answer\}}

\textbf{reasoning:} Explain whether continuing was necessary for error recovery. \\
\textbf{erc\_correct:} Answer 'yes' if continuation was the correct action to avoid 
submitting an incorrect answer, otherwise 'no'.
\end{tcolorbox}

\subsection{COMPASS Agent Prompts}

We include here the high-level instructions provided to each component in COMPASS. 
These prompts implement the architectural separation introduced in Section~\ref{sec:framework}: 
a Main Agent for tactical execution, a Context Manager for maintaining and synthesizing context, 
a Meta-Thinker for asynchronous strategic oversight, and an Answer Synthesizer for producing 
the final output. (Note that these prompts are for demonstration purpose, and the exact prompts should adapt to different benchmarks as needed).

\begin{tcolorbox}[colback=blue!5, colframe=blue!75!black, breakable, title=Main Agent Prompt]
You are the Main Agent. Your role is to execute the user's task through an 
iterative loop of reasoning, tool use, and observation. At each step you must:

\begin{enumerate}
\item Read the current task context provided to you.
\item Decide on one action or tool call to perform.
\item Execute only one tool at a time (e.g., search, retrieve, write, verify).
\item Observe the result and update your reasoning.
\item Repeat until you believe the task is complete.
\end{enumerate}

\textbf{Guidelines:}
\begin{itemize}
\item Focus on tactical execution using the immediate, curated context provided.
\item Do not attempt to monitor or evaluate global progress yourself.
\item Be explicit in reasoning: explain why the chosen tool is relevant.
\item Stop execution and return your final answer when you believe the task 
  requirements are satisfied.
\end{itemize}
\end{tcolorbox}

\begin{tcolorbox}[colback=orange!5, colframe=orange!75!black, breakable, title=Meta-Thinker Prompt]
You are the Meta-Thinker. You run asynchronously in parallel to the Main Agent. 
Your job is to monitor execution for strategic anomalies and completion signals. 

\textbf{Your tasks:}
\begin{enumerate}
\item Continuously observe the Main Agent's actions and outputs.
\item Detect anomalies such as repeated failures, contradictions, or wasted effort.
\item Detect signals that suggest the task may be complete.
\item When triggered, decide whether to:
   \begin{itemize}
   \item Persist (allow the Main Agent to continue).
   \item Pivot (redirect strategy).
   \item Verify (pause and request additional checks).
   \item Terminate (stop execution and return the final answer).
   \end{itemize}
\end{enumerate}

\textbf{Guidelines:}
\begin{itemize}
\item Your monitoring is lightweight and only activates when necessary.
\item Issue intervention or stopping signals explicitly when you detect risk of 
  compounding errors or when the answer is already sufficient.
\item Do not duplicate the Main Agent's work; focus on higher-level judgment.
\end{itemize}
\end{tcolorbox}

\begin{tcolorbox}[colback=green!5, colframe=green!75!black, breakable, title=Context Manager Prompt]
\label{app:contextmgr}
\textbf{Role.} You are the \textbf{Context Manager}. You transform full task history and research notes into a concise, execution-ready \emph{context} for the Main Agent each turn.

\textbf{Inputs (provided to you):}
\begin{itemize}
  \item \textbf{Original Question} $q$.
  \item \textbf{Previous Notes} $n_t$: structured briefs from earlier rounds (high-level summaries, verified evidence, constraints).
  \item \textbf{Current Research Progress} $\mathcal{T}_t$: the turn-level trace (key observations, tool results, reflections).
  \item \textbf{Meta Decision} \textit{decision}: \textsc{continue}/\textsc{reflect}/\textsc{replan}/\textsc{verify}/\textsc{stop}.
\end{itemize}

\textbf{Your Single Output:} produce a \emph{structured context} $x_t$ for the Main Agent (concise, specific, and directly actionable). Use the following sections:
\begin{enumerate}
  \item \textbf{Task}: one-sentence restatement of $q$.
  \item \textbf{Most-Recent Evidence}: 2--5 bullet points of verified, relevant facts (cite tool names/sources inline, if applicable).
  \item \textbf{Critical Constraints \& Corrections}: formatting/grounding constraints (e.g., “must cite FDA source”), and any corrections to earlier mistakes.
  \item \textbf{Open Items}: unresolved sub-questions or missing data (prioritized).
  \item \textbf{Next Actions (Plan)}: 2--4 concrete steps aligned with \textit{decision} (tools to call, targets, success criteria).
  \item \textbf{Tool Hints (Optional)}: specific tools to use
\end{enumerate}

\textbf{Guidelines:}
\begin{itemize}
  \item Be strictly selective: include only information required for the next turn’s tactical reasoning.
  \item Do not execute tasks; do not duplicate raw history—promote only salient facts and decisions.
  \item Keep the output compact (typically \(\leq\) 200--300 tokens), with bullet lists over prose when possible.
\end{itemize}
\end{tcolorbox}

\begin{tcolorbox}[colback=purple!5, colframe=purple!75!black, breakable, title=Answer Synthesizer Prompt]
\label{app:answersynth}
\textbf{Role.} You are the \textbf{Answer Synthesizer}. When the Meta-Thinker signals completion, you generate the final user-facing answer.

\textbf{Inputs:}
\begin{itemize}
  \item Final trace  (key observations and reasoning).
  \item Notes $n_t$ (structured briefs from prior rounds).
  \item Original question $q$ and any explicit constraints.
\end{itemize}

\textbf{Your Tasks:}
\begin{enumerate}
  \item Integrate evidence from $\mathcal{T}_t$ and $n_t$.
  \item Prioritize verified, high-confidence findings and resolve minor inconsistencies.
  \item Produce a clear, direct answer to $q$.
\end{enumerate}

\textbf{Guidelines:}
\begin{itemize}
  \item Use concise, authoritative natural language.
  \item If needed, include a one-line justification citing key evidence sources.
  \item Avoid hedging such as “insufficient information”; provide your strongest synthesis.
\end{itemize}
\end{tcolorbox}

\begin{tcolorbox}[colback=blue!4, colframe=blue!65!black, breakable, title=Note Append (Pseudocode)]
\label{app:notestore}
\lstset{language={},basicstyle=\ttfamily\footnotesize,breaklines=true,columns=flexible}
\begin{lstlisting}
# Inputs:
#   turn_id       : integer round index t
#   context_text  : structured output from Context Manager (Sections 1-6)
# Output:
#   note store updated with sections 2-4 for round t


function AppendNote(turn_id, context_text):
    # 1) Parse only the Context Manager sections we keep as notes:
    #    2) Most-Recent Evidence
    #    3) Critical Constraints & Corrections
    #    4) Open Items
    evidence    <- ExtractBullets(context_text, header="Most-Recent Evidence")
    constraints <- ExtractBullets(context_text, header="Critical Constraints & Corrections")
    open_items  <- ExtractBullets(context_text, header="Open Items")

    # 2) Form a compact record tagged by round:
    record <- {
        round: turn_id,
        evidence: evidence,
        constraints: constraints,
        open_items: open_items
    }

    # 3) Append to the rolling note store and persist:
    NoteStore.Append(record)
    NoteStore.Save()         # implementation-specific persistence (e.g., InMemory)

# Helper: returns a list of bullet lines under a given section header.
function ExtractBullets(text, header):
    # Locate 'header:' line (with optional numbering like "2) header:")
    # Collect subsequent bullet/numbered lines until next top-level header.
    bullets <- []
    ...      # implementation-specific parsing
    return bullets
\end{lstlisting}
\end{tcolorbox}

\section{Resources, Data, and Training}
\label{app:resources}

Here is the inner loop, Re-act style alogirhitm:

\begin{algorithm}[t]
\caption{ExecuteTurn: Inner Loop of Tactical Reasoning.}
\label{alg:inner}
\begin{algorithmic}[1]
\Require $\mathcal{A}^{\text{main}}, \mathcal{A}^{\text{meta}}$, \emph{context} $x$, max inner iterations $I_{\max}$
\Ensure Partial trace $\mathcal{T}$
\State $\mathcal{T}\gets[]$
\State $\mathcal{A}^{\text{meta}}.\text{StartMonitoring}(\mathcal{T})$ \Comment{Begin async monitoring of trace queue}
\For{$i = 0, 1, \ldots, I_{\max}-1$}
  \If{$\mathcal{A}^{\text{meta}}.\text{IsTriggered}()$ \textbf{or} $\mathcal{A}^{\text{main}}.\text{HasFinalAnswer}()$}
    \State \textbf{break} \Comment{Exit on anomaly or completion}
  \EndIf
  \State $\textit{step} \gets \mathcal{A}^{\text{main}}.\text{Execute}(x, \mathcal{T}, \mathcal{O})$
  \State $\mathcal{T}.\text{append}(\textit{step})$ \Comment{Meta-agent observes queue changes asynchronously}
\EndFor
\State $\mathcal{A}^{\text{meta}}.\text{StopMonitoring}()$ \Comment{End async monitoring}
\State \Return $\mathcal{T}$
\end{algorithmic}
\end{algorithm}

\subsection{Benchmarks}
We evaluate COMPASS on benchmarks that stress long-horizon reasoning across multiple interactions. Short-form QA datasets such as \textbf{SimpleQA} and \textbf{GPQA} were piloted but omitted, since strong base models already achieve near-saturation and these tasks do not benefit from tool augmentation. Our main evaluation therefore focuses on \textbf{DeepResearch-relevant benchmarks} that typically require more than 20 reasoning–action steps:
\begin{itemize}
    \item \textbf{GAIA} \citep{mialon2024gaia}: all Level 1–3 tasks without images, spanning diverse scientific and commonsense domains.
    \item \textbf{BrowseComp} \citep{wei2025browsecompsimplechallengingbenchmark}: 1{,}266 questions requiring sustained web navigation, cross-source verification, and entangled fact retrieval with short, verifiable answers.
    \item \textbf{Humans’ Last Exam (HLE)} \citep{phan2025hlexam}: 2{,}158 questions across mathematics, humanities, and natural sciences, excluding image-based or non-tool-relevant cases.
\end{itemize}
These benchmarks collectively cover diverse failure modes in long-horizon reasoning: cascading search errors, tool API misuses, and premature stopping.

\subsection{Baselines and Models}
We compare against two categories.  
\textbf{Fundamental paradigms}: single-agent (\emph{Search/Browse}, \emph{+Meta-Thinking}, \emph{+Context}) and multi-agent (\emph{Agent-as-a-Tool}, \emph{Decentralized Handoffs}, \emph{Plan-and-Execute} \citep{wang-etal-2023-plan}).  
\textbf{Established research agents}: OpenAI \emph{DeepResearch}, DeepSeek Agent, and Google \emph{TestTime Diffusion}. For fair comparison, test-time scaling (parallel sampling) is included only when benchmarking against established systems.  
All experiments use \textbf{Gemini 2.5 Pro} and \textbf{Gemini 2.5 Flash} as backbone reasoning models, with \textbf{Gemma-3-12B} in ablations for specialized Context Manager training. For reproducibility, we fix random seeds across sampling runs and log all trajectories for post-hoc auditing.

\subsection{Evaluation Metrics}
Our primary metric is \textbf{accuracy (Pass@1)} defined per benchmark-specific criteria. We additionally measure \textbf{token usage} (both per-step and end-to-end) to capture efficiency. To assess strategic reliability, we introduce four \textbf{meta-thinking metrics} (PAR, PVR, CA, ERC) as formalized in Section~\ref{sec:framework}, each judged via LLM-as-a-Judge with structured outputs. See Appendix~\ref{app:prompts} for full grading prompts. We report mean values over three runs to mitigate stochasticity from sampling and tool call variability.

\subsection{Synthetic Data and Training for Context Manager}
\label{app:training}

We construct training data by mining trajectories from GAIA, SimpleQA, HotPotQA, and GPQA, MMLU or MMLU-Pro, in order to cover different reasoning scenarios, using the COMPASS inference pipeline. From 13{,}486 raw trajectories, we filter down to \textbf{2{,}065} high-quality examples emphasizing (i) complete task solutions, (ii) error-recovery sequences, and (iii) proper termination after intermediate success. We exclude trivial paths ($<3$ tool calls) and degenerate answers without reflection, and we upsample recovery cases to improve robustness. Each example pairs trajectory history and Meta-Thinker reflections with the optimized context for the next turn.

\subsubsection{Data Preprocessing and Quality Control}

Our data construction pipeline applies several quality filters to ensure training examples capture the structured reasoning patterns required for context management:

\textbf{Trajectory Length Filtering:} We retain only trajectories with 3-25 tool calls, excluding both trivial single-step solutions and excessively long sequences that may contain repetitive failures.

\textbf{Success Pattern Analysis:} Training examples are categorized into three types: (i) \textit{direct success} trajectories that solve tasks without errors, (ii) \textit{recovery sequences} where agents overcome initial mistakes through strategic pivots, and (iii) \textit{verification patterns} where agents validate solutions before concluding. We upsample recovery sequences (2.3× multiplier) to improve the model's ability to synthesize context after failures.

\textbf{Context Complexity Stratification:} We balance examples across varying context complexities: simple constraint tracking (35\%), multi-source evidence synthesis (45\%), and complex constraint interaction cases (20\%). This ensures the model learns both basic summarization and sophisticated context organization.

\subsubsection{Supervised Fine-Tuning (SFT)}
We fine-tune the model to generate concise context briefs that preserve critical constraints and strategic signals from Meta-Thinker reflections while maintaining structured output format for pipeline integration.

\textbf{Dataset Construction:} 
We curate a training dataset of 10{,}347 examples comprising:
\begin{itemize}
    \item Query-context pairs where contexts distill verbose planning queries into minimal briefs while preserving all critical constraints
    \item Contexts derived from Meta-Thinker reflections, maintaining strategic planning signals
    \item Structured outputs following our standardized template (objective restatement, progress tracking, constraint preservation, next-step context)
\end{itemize}
The dataset is constructed from COMPASS pipeline executions, with human annotation to ensure quality and template adherence.

\textbf{Training Objective:}
We use standard cross-entropy loss for autoregressive language modeling:
$$\mathcal{L}_{SFT} = -\sum_{t=1}^{T} \log P(y_t \mid y_{<t}, x; \theta)$$
where $x$ represents the input query with Meta-Thinker reflections, and $y$ is the target context brief.

\textbf{Training Setup:} 
We fine-tune \textbf{Gemma-3-12B} for 3{,}000 steps on four A100 GPUs (80GB) using DeepSpeed ZeRO Stage-2 with gradient accumulation. Hyperparameters: AdamW optimizer ($\beta_1=0.9, \beta_2=0.999$), effective batch size 32, learning rate $1\times10^{-4}$ with cosine decay to $1\times10^{-6}$, weight decay $0.05$, warmup ratio 0.1. We apply gradient clipping (max norm 1.0) and monitor validation perplexity every 200 steps, selecting the checkpoint with lowest perplexity on a held-out validation set ($N=500$).

\textbf{Template Consistency:} 
All training examples follow our structured format, ensuring downstream compatibility with the COMPASS pipeline. We validate format compliance through post-processing checks, achieving 94.3\% adherence on held-out test data.

\subsubsection{Direct Preference Optimization (DPO)}
Following similar procedure as the SFT above, we apply DPO~\citep{rafailov2024direct} to further align the context generation model toward task-efficient summaries. DPO enables direct optimization from preference data without requiring explicit reward models or reinforcement learning.

\textbf{Preference Data Collection:}
For each of training queries from the above SFT dataset, we generate four candidate context summaries from Context-12B-SFT using nucleus sampling with varying temperatures $\tau \in \{0.7, 0.9, 1.1, 1.3\}$ and top-$p=0.95$. Each candidate is evaluated by executing the full COMPASS pipeline and measuring:
\begin{itemize}
    \item Task success (binary completion indicator)
    \item Token efficiency (total tokens consumed until completion or timeout)
    \item Strategic appropriateness (composite score from PAR/PVR/CA/ERC metrics)
\end{itemize}

\textbf{Preference Pair Construction:}
We rank the four candidates per trajectory by a composite metric: $\text{score} = \mathbf{1}_{\{\text{success}\}} - 0.001\,\text{tokens}$
, where the token penalty encourages efficiency without sacrificing success. We construct preference pairs $(y_w, y_l)$ from adjacent rankings (rank $i$ vs. rank $i+1$), yielding up to 6 pairs per trajectory.

To ensure meaningful preferences, we filter pairs where: (1) both contexts lead to failure, (2) the success rate difference is $<5\%$ on our validation set, or (3) the preferred context uses $>20\%$ more tokens without measurable quality improvement. This filtering yields 8{,}200 high-quality preference pairs.

\textbf{Training Setup:}
We optimize the DPO objective:
\begin{align*}
\mathcal{L}_{\text{DPO}}(\theta) = -\mathbb{E}_{(x,y_w,y_l)}\bigg[\log\sigma\bigg(\beta \log\frac{\pi_\theta(y_w|x)}{\pi_{\text{ref}}(y_w|x)} \\
- \beta\log\frac{\pi_\theta(y_l|x)}{\pi_{\text{ref}}(y_l|x)}\bigg)\bigg]
\end{align*}
where $\pi_{\text{ref}}$ is Context-12B-SFT (frozen), $\pi_\theta$ is the policy being optimized, and $\beta=0.1$ controls the KL penalty strength.

Training runs for 6{,}000 steps on eight A100 GPUs with DeepSpeed ZeRO Stage-2. Hyperparameters: AdamW optimizer, batch size 32 (4 per GPU), learning rate $5\times10^{-5}$ with linear warmup (10\% of steps) and cosine decay, gradient clipping (max norm 1.0). We evaluate on a held-out validation set of 200 trajectories every 500 steps and select the checkpoint maximizing validation success rate.

\textbf{Results:}
The final DPO model (Context-12B-DPO) achieves 30\% token reduction compared to Context-12B-SFT (from 2{,}847 to 1{,}993 average tokens per task) while maintaining comparable task success rates (87.3\% vs. 87.8\% on our test set). Notably, the DPO model shows improved strategic appropriateness scores, suggesting better alignment with efficient planning trajectories.

\subsection{Resources and Infrastructure}

All Gemini API calls and long-term memory operations were executed through services and infrastructure provide by google cloud infrastructure, leveraging its managed memory bank services for trajectory storage and retrieval. Training was performed with up to 8 A100-80GB GPUs and huggingface's trl library was used for SFT and DPO implementations. We use Agents Development Kit (ADK) library to implement multi-agent orchestration. To ensure reproducibility, we log tool API events, token counts, and intermediate contexts; all experiments were run with fixed seeds and capped API retries to handle stochastic tool failures.

\textbf{Computational Requirements:} Total training compute for Context-12B required approximately 480 GPU-hours across SFT and DPO phases. Data preprocessing and preference pair construction consumed an additional 120 CPU-hours. All training runs used mixed precision (fp16) with gradient checkpointing to manage memory usage efficiently.

\section{Case Studies}
\label{app:case_all}

\newcommand{\tinybadge}[2]{%
  \tikz[baseline=(char.base)]{%
    \node[shape=rectangle, rounded corners=2pt, inner sep=2pt, 
          fill=#1!20, draw=#1!50!black, text=#1!50!black, 
          font=\tiny\sffamily] (char) {\sloppy #2};%
  }%
}

\begin{figure*}[t]
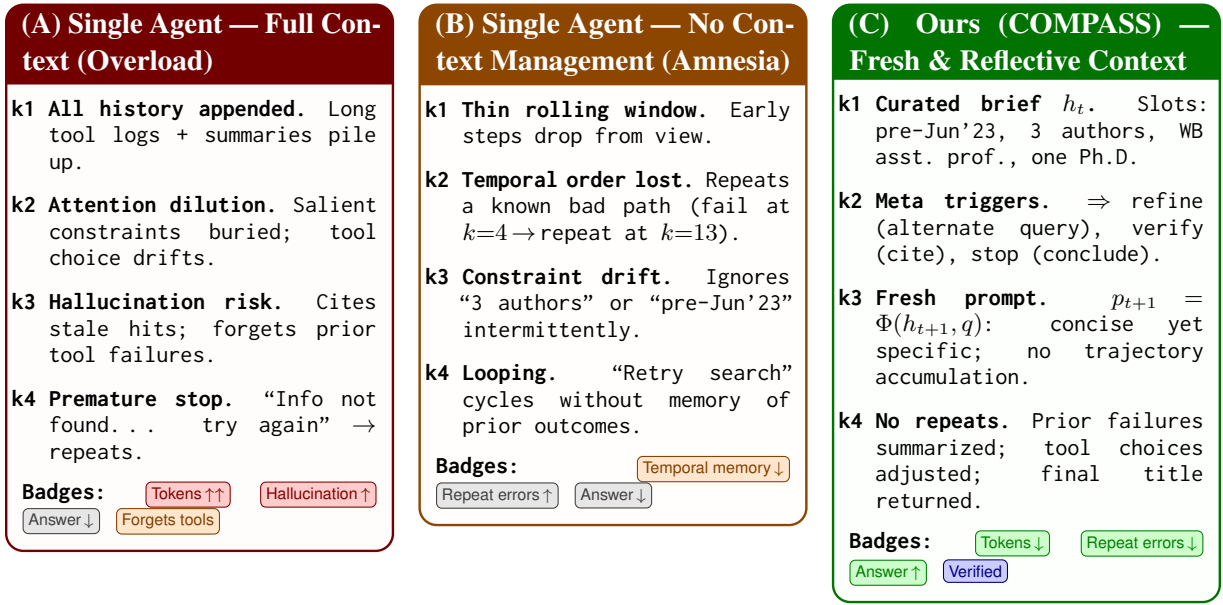
 
\centering
\begin{minipage}[t]{0.32\linewidth}
\begin{tcolorbox}[colback=red!1,colframe=red!45!black,boxrule=0.9pt,arc=6pt,
title=\textbf{(A) Single Agent — Full Context (Overload)}]
\footnotesize
\begin{enumerate}[leftmargin=1.1em,itemsep=2pt,topsep=2pt,label=\textbf{k\arabic*}]
  \item \textbf{All history appended.} Long tool logs + summaries pile up.
  \item \textbf{Attention dilution.} Salient constraints buried; tool choice drifts.
  \item \textbf{Hallucination risk.} Cites stale hits; \emph{forgets} prior tool failures.
  \item \textbf{Premature stop.} “Info not found… try again” \(\rightarrow\) repeats.
\end{enumerate}
\vspace{2pt}
\textbf{Badges:} \tinybadge{red}{Tokens\,\(\uparrow\uparrow\)}\,
\tinybadge{red}{Hallucination\,\(\uparrow\)}\,
\tinybadge{gray}{Answer\,\(\downarrow\)}\,
\tinybadge{orange}{Forgets tools}
\end{tcolorbox}
\end{minipage}
\hfill
\begin{minipage}[t]{0.32\linewidth}
\begin{tcolorbox}[colback=orange!1,colframe=orange!55!black,boxrule=0.9pt,arc=6pt,
title=\textbf{(B) Single Agent — No Context Management (Amnesia)}]
\footnotesize
\begin{enumerate}[leftmargin=1.1em,itemsep=2pt,topsep=2pt,label=\textbf{k\arabic*}]
  \item \textbf{Thin rolling window.} Early steps drop from view.
  \item \textbf{Temporal order lost.} Repeats a known bad path (fail at \(k{=}4\)\,\(\rightarrow\)\,repeat at \(k{=}13\)).
  \item \textbf{Constraint drift.} Ignores “3 authors” or “pre-Jun’23” intermittently.
  \item \textbf{Looping.} “Retry search” cycles without memory of prior outcomes.
\end{enumerate}
\vspace{2pt}
\textbf{Badges:} \tinybadge{orange}{Temporal memory\,\(\downarrow\)}\,
\tinybadge{gray}{Repeat errors\,\(\uparrow\)}\,
\tinybadge{gray}{Answer\,\(\downarrow\)}
\end{tcolorbox}
\end{minipage}
\hfill
\begin{minipage}[t]{0.32\linewidth}
\begin{tcolorbox}[colback=green!1,colframe=green!45!black,boxrule=0.9pt,arc=6pt,
title=\textbf{(C) \;Ours (COMPASS) — Fresh \& Reflective Context}] 
\footnotesize
\begin{enumerate}[leftmargin=1.1em,itemsep=2pt,topsep=2pt,label=\textbf{k\arabic*}]
  \item \textbf{Curated brief \(h_t\).} Slots: \emph{pre-Jun’23}, \emph{3 authors}, \emph{WB asst.\ prof.}, \emph{one Ph.D.}
  \item \textbf{Meta triggers.} \(\Rightarrow\) \textsc{refine} (alternate query), \textsc{verify} (cite), \textsc{stop} (conclude).
  \item \textbf{Fresh prompt.} \(p_{t+1}=\Phi(h_{t+1},q)\): concise yet specific; no trajectory accumulation.
  \item \textbf{No repeats.} Prior failures summarized; tool choices adjusted; final title returned.
\end{enumerate}
\vspace{2pt}
\textbf{Badges:} \tinybadge{green}{Tokens\,\(\downarrow\)}\,
\tinybadge{green}{Repeat errors\,\(\downarrow\)}\,
\tinybadge{green}{Answer\,\(\uparrow\)}\,
\tinybadge{blue}{Verified}
\end{tcolorbox}
\end{minipage}
\caption{Comparison of agent designs across context management strategies.}
\label{fig:agent_comparison}
\end{figure*}

This section provides detailed trajectory examples illustrating the meta-thinking decision scenarios outlined in Table~\ref{tab:meta-thinking}. We use the following complex query from BrowseComp to demonstrate failure patterns in long-horizon tasks:

\textbf{Query:} Between 1990 and 1994 inclusive, what teams played in a soccer match with a Brazilian referee had four yellow cards, two for each team where three of the total four were not issued during the first half, and four substitutions, one of which was for an injury in the first 25 minutes of the match.

\textbf{Ground Truth Answer:} Ireland v Romania

\subsection{Scenario 1A: Local Error - Correct Persistence}
\label{app:local_error_persist}

This trajectory demonstrates appropriate persistence when facing a local search refinement need.

\begin{lstlisting}[style=trajectory, caption=Trajectory showing correct persistence through local error]
Step 1: <think>
Need to find soccer match 1990-1994 with specific yellow card and substitution patterns with Brazilian referee.
</think>
<tool_call name="search">
soccer match 1990-1994 Brazilian referee yellow cards substitutions
</tool_call>
<tool_response>
Results show various World Cup matches from 1990-1994, but no specific yellow card details visible in snippets.
</tool_response>

Step 2: <think>
Search too broad. Need to focus on World Cup 1990 and 1994 specifically since those had Brazilian referees.
</think>
<tool_call name="search">
World Cup 1990 1994 Brazilian referee matches yellow cards
</tool_call>
<tool_response>
Found several matches with Brazilian referees. Notable: Ireland vs Romania 1990 with referee Wright (not Brazilian). Need to refine.
</tool_response>

Step 3: <think>
Still not finding Brazilian referee specifically. Let me search for Brazilian FIFA referees active 1990-1994.
</think>
<tool_call name="search">
Brazilian FIFA referee World Cup 1990 1994 official matches
</tool_call>
<tool_response>
Found: Jose Wright was actually Jose Roberto Wright, Brazilian referee who officiated Ireland vs Romania, Italy 1990.
</tool_response>
\end{lstlisting}

\textbf{Meta-Decision:} CONTINUE - The agent correctly identifies that the search strategy is fundamentally sound but needs refinement. Each step builds logically on the previous, gradually narrowing the search scope.

\subsection{Scenario 1B: Global Dead-End - Correct Revision}
\label{app:global_deadend_revise}

This trajectory shows appropriate strategic revision when the current approach hits a fundamental barrier.

\begin{lstlisting}[style=trajectory, caption=Trajectory requiring strategic revision]
Step 1-4: [Similar initial searches as above]

Step 5: <think>
Found Jose Roberto Wright as Brazilian referee, but need specific yellow card and substitution details for Ireland vs Romania match.
</think>
<tool_call name="search">
Ireland Romania 1990 World Cup yellow cards substitutions details
</tool_call>
<tool_response>
General match reports available, but detailed yellow card timing and substitution reasons not in search results.
</tool_response>

Step 6-8: [Repeated attempts with similar queries yield no detailed card/substitution timing]

Step 9: <think>
Search engines not providing the granular match details needed. Need to try FIFA official records or match reports.
</think>
<tool_call name="search">
site:fifa.com Ireland Romania 1990 World Cup match report official
</tool_call>
<tool_response>
FIFA site search limited. No detailed match reports available in search results.
</tool_response>

Step 10: <think>
Current web search approach hitting dead end. Need to pivot to historical sports databases or newspaper archives from 1990.
</think>
<tool_call name="search">
newspaper archive 1990 Ireland Romania World Cup match report yellow cards
</tool_call>
<tool_response>
Found detailed match report from Irish Times archive: confirms 4 yellow cards (2 each team), 3 in second half, 4 substitutions including injury substitution in 23rd minute.
</tool_response>
\end{lstlisting}

\textbf{Meta-Decision:} REVISE - After steps 6-9 showed repeated failure to find detailed match statistics through general web search, the agent correctly pivoted to historical newspaper archives, which contained the required granular information.

\subsection{Scenario 2A: Correct Solution - Appropriate Conclusion}
\label{app:correct_conclude}

This trajectory demonstrates proper recognition of task completion.

\begin{lstlisting}[style=trajectory, caption=Trajectory with correct conclusion timing]
Step 1-7: [Successful search progression leading to match identification]

Step 8: <think>
Found comprehensive evidence: Ireland vs Romania, 1990 World Cup, Brazilian referee Jose Roberto Wright, 4 yellow cards (2 each team, 3 in 2nd half), 4 substitutions including injury sub in 23rd minute. All criteria satisfied.
</think>
<tool_call name="verify">
Cross-check: Ireland Romania 1990 World Cup referee Jose Roberto Wright Brazilian
</tool_call>
<tool_response>
Confirmed: Jose Roberto Wright, Brazilian referee, officiated Ireland 1-0 Romania, Italy 1990 World Cup.
</tool_response>

Step 9: <think>
All query requirements verified. Have definitive answer with supporting evidence.
</think>
<answer>Ireland v Romania</answer>
\end{lstlisting}

\textbf{Meta-Decision:} CONCLUDE - The agent appropriately recognizes that all query constraints have been satisfied and verified, leading to correct termination.

\subsection{Scenario 2B: Incorrect Solution - Necessary Continuation}
\label{app:incorrect_continue}

This trajectory illustrates the importance of continued verification when initial conclusions are flawed.

\begin{lstlisting}[style=trajectory, caption=Trajectory requiring error recovery through continuation]
Step 1-5: [Initial searches identify several 1990-1994 matches with Brazilian referees]

Step 6: <think>
Found Brazil vs Argentina 1990 with Brazilian referee. This seems like a clear match - Brazilian referee in a Brazil game.
</think>
<tool_call name="search">
Brazil Argentina 1990 World Cup yellow cards substitutions
</tool_call>
<tool_response>
Match report shows 6 yellow cards total, not 4. Also only 2 substitutions, not 4 required.
</tool_response>

Step 7: <think>
Brazil vs Argentina doesn't match the constraints. But wait - would a Brazilian referee officiate a Brazil match? That seems unlikely for neutrality.
</think>
<tool_call name="search">
FIFA referee neutrality rules Brazilian referee Brazil matches World Cup
</tool_call>
<tool_response>
FIFA rules require neutral referees. Brazilian referees cannot officiate Brazil national team matches.
</tool_response>

Step 8: <think>
My initial assumption was wrong. Need to find matches between non-Brazilian teams with Brazilian referees.
</think>
<tool_call name="search">
1990 World Cup Brazilian referee non-Brazil matches yellow cards
</tool_call>
<tool_response>
Jose Roberto Wright (Brazilian) officiated Ireland vs Romania, among others.
</tool_response>
\end{lstlisting}

\textbf{Meta-Decision:} CONTINUE - The agent initially reached an incorrect conclusion (Brazil vs Argentina) but through continued analysis discovered the logical flaw in having a Brazilian referee officiate a Brazil match, leading to the correct answer.

\subsection{Failure Pattern Analysis}

These trajectories illustrate the two major failure patterns identified in our research:

\textbf{Context Overflow:} In scenarios without proper context management, agents accumulate extensive search histories that bury critical constraints (e.g., the "Brazilian referee" requirement becomes lost among dozens of search results about various World Cup matches).

\textbf{Premature Termination:} Agents may conclude with superficially plausible but incorrect answers (e.g., Brazil vs Argentina because it involves a Brazilian referee) without proper verification of all constraints.

The COMPASS framework addresses these through explicit meta-monitoring that can trigger strategic revisions (Scenario 1B) or prevent premature conclusions (Scenario 2B) while maintaining efficiency through appropriate continuation decisions (Scenarios 1A and 2A).

\subsection{Agent Complexity Comparison}
\label{app:case_study_comparison}

To illustrate the varying cognitive demands across COMPASS components, we present parallel outputs from all three agents during a single trajectory step. This analysis informed our decision to develop a specialized compact Context Manager.

\textbf{Query:} Find the exact publication date of the most recent FDA guidance on machine learning in medical devices.

\textbf{Current Step:} Agent has performed 8 search attempts, found multiple FDA documents from different years, but discovered conflicting publication dates across sources.

\subsubsection{Main Agent Output}
The Main Agent performs tactical reasoning with tool selection:

\begin{lstlisting}[style=trajectory, caption=Main Agent tactical execution]
<think>
Found FDA documents from 2021, 2022, and 2023, but dates are inconsistent across sources. 
The 2023 document mentions "updated guidance" but some sources show 2022 as most recent.
Need to verify directly from FDA website to get authoritative publication date.
Should search specifically on FDA.gov domain to avoid third-party date errors.
</think>
<tool_call name="search">
site:fda.gov "machine learning" "medical devices" guidance 2023 publication date
</tool_call>
\end{lstlisting}

\textbf{Analysis:} The Main Agent focuses on immediate next steps, tool selection, and tactical problem-solving. Its reasoning is linear and execution-focused.

\subsubsection{Meta-Thinker Output}
The Meta-Thinker performs strategic oversight and anomaly detection:

\begin{lstlisting}[style=trajectory, caption=Meta-Thinker strategic analysis]
<analysis>
TRAJECTORY ASSESSMENT:
- 8 search attempts completed
- Multiple candidate documents identified (2021, 2022, 2023)
- ANOMALY DETECTED: Date inconsistencies across sources suggest information quality issues
- RISK: Agent may conclude with incorrect date due to conflicting evidence

STRATEGIC EVALUATION:
Current approach: Appropriate focus on authoritative FDA source
Persistence vs Pivot: CONTINUE - strategy is sound, execution refining toward official source
Completion readiness: NOT READY - conflicting evidence requires verification

INTERVENTION DECISION: VERIFY
Rationale: Before concluding, agent should cross-reference publication metadata 
from official FDA document rather than relying on search result summaries.

MONITORING PRIORITY: High - date accuracy critical for regulatory information
</analysis>
\end{lstlisting}

\textbf{Analysis:} The Meta-Thinker engages in complex strategic reasoning, anomaly detection, risk assessment, and multi-level decision making about trajectory management.

\subsubsection{Context Manager Output}
The Context Manager performs structured summarization and context synthesis:

\begin{lstlisting}[style=trajectory, caption=Context Manager context synthesis]
UPDATED CONTEXT BRIEF:

OBJECTIVE: Find exact publication date of most recent FDA ML medical device guidance

SEARCH PROGRESS:
- Completed: 8 search iterations
- Sources explored: FDA documents, medical journals, regulatory news sites
- Candidate documents identified: 
  * 2021: "Software as Medical Device" guidance
  * 2022: "Machine Learning-Enabled Medical Devices" draft
  * 2023: "Updated ML/AI Medical Device Guidance" (date disputed)

CURRENT FINDINGS:
 Confirmed: FDA has published ML medical device guidance
 Unresolved: Exact publication date of most recent version (2022 vs 2023 conflict)

KEY CONSTRAINTS:
- Must be official FDA publication (not draft or third-party summary)
- Need authoritative publication date, not revision or comment period dates
- Focus on "medical devices" specifically (not general AI/ML guidance)

NEXT STEP CONTEXT:
Agent attempting FDA.gov domain search to resolve date conflict.
Priority: Verify official publication metadata from primary source.
\end{lstlisting}

\textbf{Analysis:} The Context Manager performs systematic information organization, constraint tracking, and structured summarization. Its output follows predictable templates and focuses on factual synthesis rather than strategic reasoning.


\end{document}